\pdfoutput=1

\documentclass[11pt]{article}

\usepackage{ACL2023}

\usepackage{times}
\usepackage{latexsym}

\usepackage[T1]{fontenc}

\usepackage[utf8]{inputenc}

\usepackage{microtype}

\usepackage{inconsolata}

\usepackage{dblfloatfix}
\usepackage{graphicx}
\usepackage{placeins}

\usepackage{listings}

\usepackage{xcolor}

\title{Automatic Glossary of Clinical Terminology: a Large-Scale Dictionary of \\
Biomedical Definitions Generated from Ontological Knowledge}

\author{François Remy \and Thomas Demeester \\
  IDLab (Internet and Data Science Lab), Ghent University \& imec \\
  \texttt{francois.remy@ugent.be}}

\begin{document}
\maketitle
\begin{abstract}
\textbf{Background: }More than 400,000 biomedical concepts and some of their relationships are contained in SnomedCT \citep{SnomedCT}, a comprehensive biomedical ontology. However, their concept names are not always readily interpretable by non-experts, or patients looking at their own electronic health records (EHR). Clear definitions or descriptions in understandable language are often not available. Therefore, generating human-readable definitions for biomedical concepts might help make the information they encode more accessible and understandable to a wider public.

\textbf{Objective: }In this article, we introduce the Automatic Glossary of Clinical Terminology (AGCT), a large-scale biomedical dictionary of clinical concepts generated using high-quality information extracted from the biomedical knowledge contained in SnomedCT.

\textbf{Methods: }We generate a novel definition for every SnomedCT concept, after prompting the OpenAI Turbo model, a variant of GPT 3.5, using a high-quality verbalization of the SnomedCT relationships of the to-be-defined concept. A significant subset of the generated definitions was subsequently judged by NLP researchers with biomedical expertise on 5-point scales along the following three axes: 
factuality, insight, and fluency.

\textbf{Results: }AGCT contains 422,070 computer-generated definitions for SnomedCT concepts, covering various domains such as diseases, procedures, drugs, and anatomy. The average length of the definitions is 49 words. The definitions were assigned average scores of over 4.5 out of 5 on all three axes, indicating a majority of factual, insightful, and fluent definitions.

\textbf{Conclusion: }AGCT is a novel and valuable resource for biomedical tasks that require human-readable definitions for SnomedCT concepts. It can also serve as a base for developing robust biomedical retrieval models or other applications that leverage natural language understanding of biomedical knowledge.

\end{abstract}

\section{Introduction}

\begin{figure*}
\centering
\includegraphics[width=1.0\textwidth]{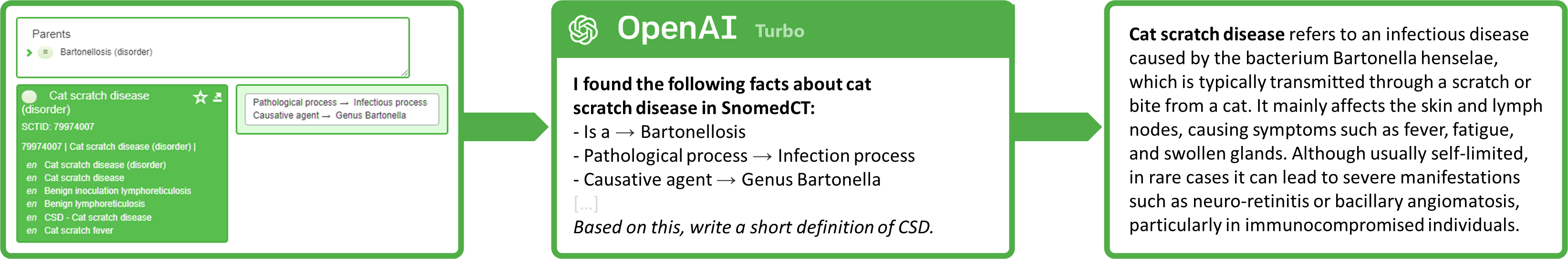}
\caption{An overview of our prompting and definition generation strategy.}
\end{figure*}

To unlock the value of in-hospital data while preserving patients’ right to privacy, federated learning might become a corner stone for retrospective studies in the healthcare domain \citep{Zerka2020SystematicRO}. Inter-hospital data interoperability is, however, one of the pre-requirements of federated learning \citep{Lamer2021SpecificationsFT} and is getting attention. 

This has resulted in a stronger appetite for more structured and more standardized coding of patient journeys through the medical services \citep{Joseph2020PatientJM}.
An issue with these standardized codes from ontologies is their usage of a highly-specialized terminology
\citep{Schulz2005SemanticCO}. This diminishes their suitability 
for non-experts or patients wishing to consult their personal clinical data. 

Efforts to produce simpler-to-understand definitions of biomedical concepts have been well-documented (e.g. by \citeauthor{MayoClinic}) but they are usually of limited scope due to the time and cost involved in their creation, requiring heavy prioritization of the efforts \citep{Jinying2017LayLanguageResources}.

Early efforts to bridge the gap between ontologies and textual definitions by \citet{Tsatsaronis2013} and \citet{Petrova2015} remained insufficient. But in recent years, the fluency of text generated using large language models has reached extremely high levels \citep{aksitov2023characterizing} and so has their ability to convert 
graph-level information into textual descriptions \citep{ribeiro2021investigating}.

In this study, 
we set out to investigate
the suitability of commercially-available language models to generate at scale medically-accurate descriptions of clinical concepts in 2023. 
To this end, we introduce the Automatic Glossary of Clinical Terminology (AGCT), a large-scale biomedical dictionary containing more than 400,000 biomedical concept definitions generated using GPT-3.5 \citep{ouyang2022training} on the basis of the biomedical knowledge contained in the SnomedCT ontology.


\section{Methodology}

Each definition was obtained after prompting the GPT-3.5 model with a long prompt containing a summarized verbalization of the SnomedCT relationships of the to-be-defined concept, as well as precise instructions for the generation of a relevant biomedical concept definition (see Annex \ref{sec:appendixB}). 

\begin{figure*}[t!]
\centering
\includegraphics[width=1.0\textwidth]{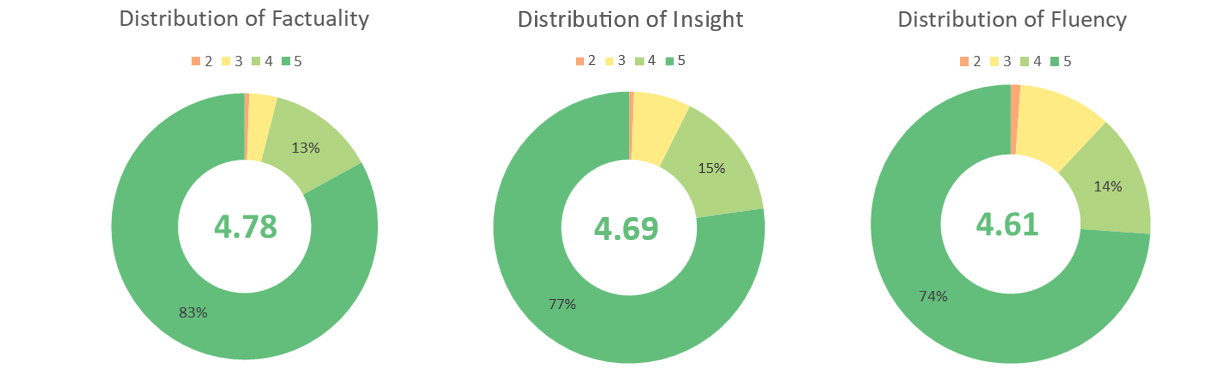}
\caption{The distribution of ratings reported by the annotators.}
\label{fig:ratings-graphs}
\end{figure*}

\begin{figure*}[t!]
\centering
\vspace{0.23cm}
\includegraphics[width=1.0\textwidth]{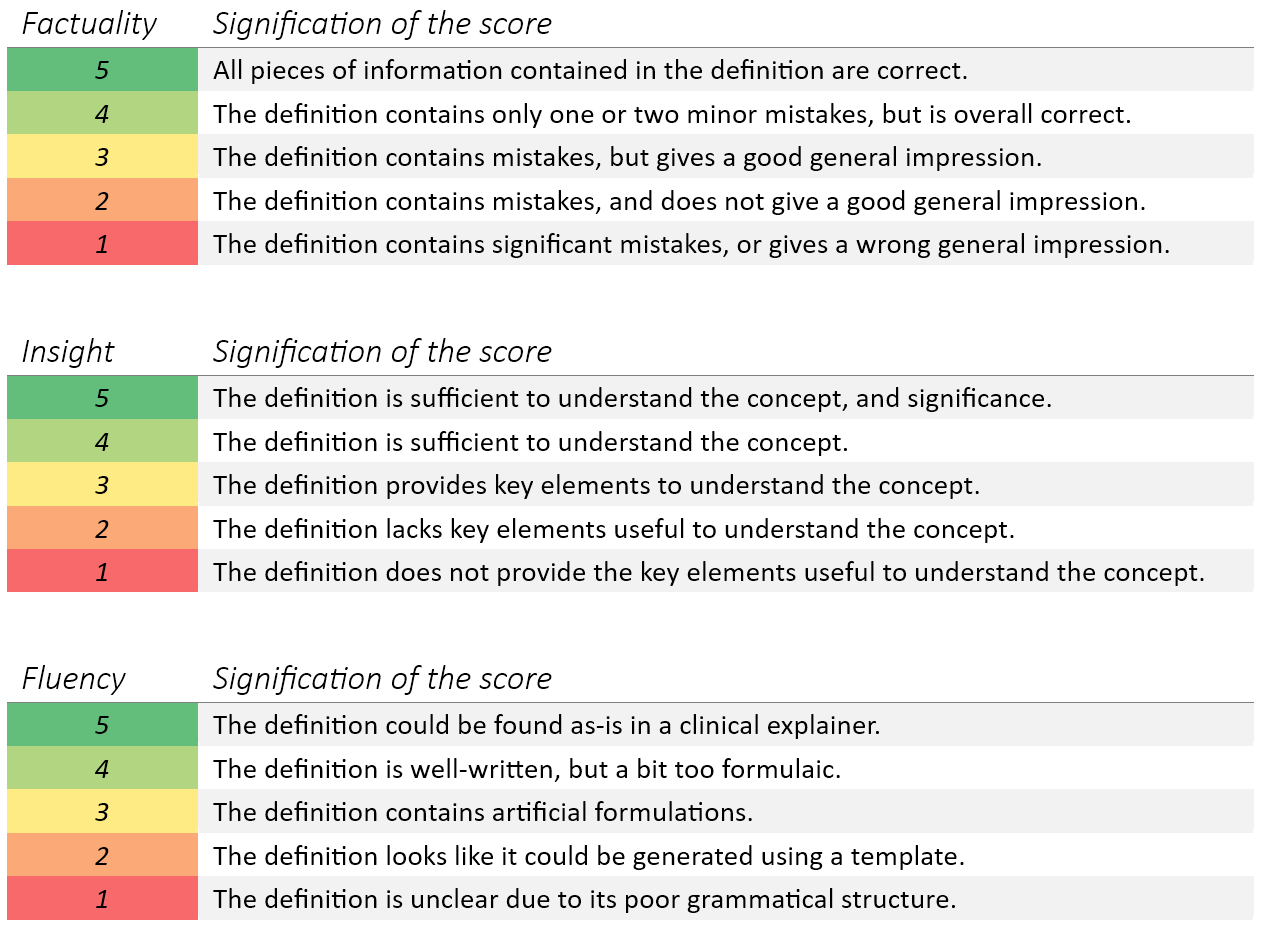}
\caption{The rating instructions, as provided to the annotators.}
\textcolor{gray}{\footnotesize{\textit{By "clinical explainer", we mean an educative document provided to patients about their condition.}}}
\label{fig:ratings-instr}
\end{figure*}
 
An important aspect of the generation process is its data curation. For every token generated by the model, four times more tokens were provided to the model in its prompt, on average. Our hypothesis was that this saturation of information should ensure that the generated definitions are well-informed and therefore factual, coupled with the already impressive biomedical knowledge of GPT-3.5 \citep{Kung2023USMLE,Aida2023USMLE}.

To confirm our intuition, 175 of the generated definitions were subsequently sent to be rated by NLP practitioners with biomedical expertise on three scales: factuality, insight, and fluency. Each of these scales is measured using a 5-point rating system, detailed further in Figure \ref{fig:ratings-instr}. 

To assess factuality, annotators relied on several Internet search queries for each definition, to verify each of the points mentioned in the definition. This was necessary as the information contained in the definition was not always readily available from a single initial search\footnote{for example, whether or not divergence paralysis was indeed related to an inability to move the eyes outwards.}.

\section{Results}

\subsection{Factuality, Insight, and Fluency}

We provide a graph representation of the resulting distributions in Figure \ref{fig:ratings-graphs}.  
The definitions generated by the proposed methodology received good ratings across all three axes measured in this study, with average scores above 4.5 for all metrics (i.e. 4.78 for Factuality, 4.69 for Insight, and 4.61 for Fluency).
For each metric taken separately, more than three quarters of the definitions obtained the highest rating on the scale.

\textbf{Factuality} turned out to be rated higher than we originally expected, with 83\% of definitions containing no incorrect information at all, and 96\% of definitions containing at most a minor mistake not hurting comprehension. It seems that the amount of information provided in the prompts, combined with GPT-3.5's world knowledge, was sufficient to prevent hallucinations in most cases.

\textbf{Insight} fared a bit worse than Factuality however, with 13\% of the definitions not containing all key elements required to properly understand the defined concept. On a positive note, only a very small fraction (<1\%) of the generated definitions lacked enough key elements to prevent the proper understanding of the defined concept.

\textbf{Fluency} fared similarly to Insight, to our surprise. In most cases where fluency was deemed lacking, the issue arose when the model tried to include less relevant or obvious details in definitions, resulting in artificial constructions that were very easily spotted by annotators. It sounds likely that fluency could have been improved by iterating on the prompt. Another common failure mode was when the prompt contained too few information, and the model wrote filler text to compensate.

\textbf{Inter-annotator agreement} was within limits for all three axes, with 
correlations around or above 0.6 for all three ratings (0.61 for factuality, 0.60 for insight, 0.58 for fluency). This is in line with the expectations for annotations by humans \citep{frost_2022}, but indicates slightly too optimistic results.

\textbf{The Pearson correlation} was low between the ratings of the three axes: among all the possible one-vs-all correlations, only one metric (Insight) turned out to be weakly correlated $(\rho = 0.40)$ with the combination of the Factuality and Fluency metrics, resulting in an ability to explain around 15\% of variability in insight using the two other variables combined $(R^2=0.16)$. All the other combinations and pairwise correlations turned out to be almost uncorrelated. This demonstrates the importance of having these three independent ratings, instead of a single rating for all definitions.


\begin{figure*}[b!]
\centering
\includegraphics[width=1.0\textwidth]{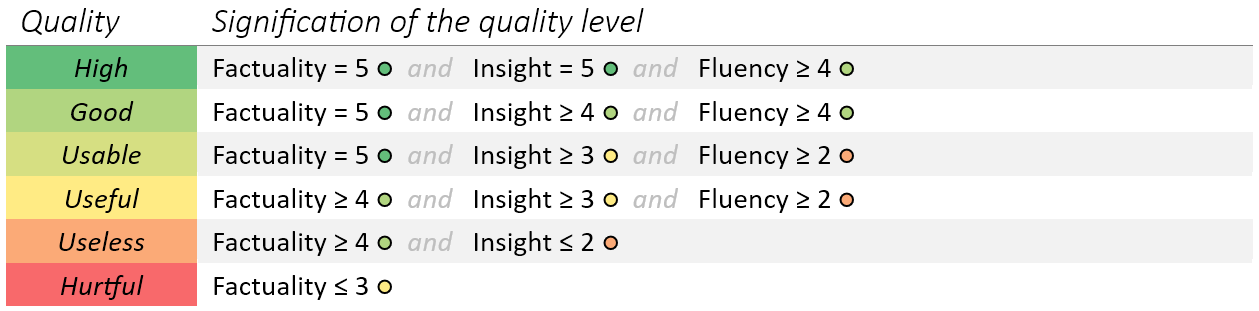}
\caption{The rules for classifying definitions into quality levels.}
\label{fig:quality-instr}
\end{figure*}

\subsection{Overall quality level}

As a result of the lack of correlation between the three variables measured by our annotators, and of their differing importance, measuring the quality of a generated definition cannot be done by simple linear combination of the ratings.

We developed a 6-levels quality scale taking into account the criticality of factuality and the lower importance of fluency, as detailed in Figure~\ref{fig:quality-instr}. Among the six levels, three levels of particular importance warrant further calling out: 

\textbf{Usable definitions} correspond to all definitions which might be presented as-is to a patient in order to help the comprehension of their EHR. These definitions need a perfect factuality rating (5) and a decent level of insight (3 or more). Usable definitions are represented by green colors.

\textbf{Useful definitions} correspond to all definitions which might be relevant for machine learning models, and in particular during the training of retrieval models for the biomedical domain. To the contrary of usable definitions, useful definitions might contain minor mistakes as long as they don’t hurt the comprehension of the concept. Useful definitions which are not also more generally usable are represented in yellow.

\textbf{Hurtful definitions} correspond to all definitions which would not be relevant for machine learning models, due to a too low level of factuality (3 or below). While some definitions might not be useful or fluent at all, as long as they remain correct, machine learning models are unlikely to be mislead by them. Definitions which contain multiple minor mistakes or major mistakes might result in incorrect results however and are thus considered hurtful for our purposes. Hurtful definitions are represented in red.

\begin{figure}[t!]
\centering
\includegraphics[width=0.43\textwidth]{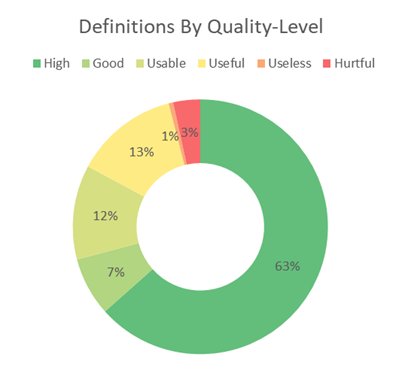}
\includegraphics[width=0.43\textwidth]{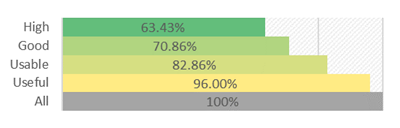}
\caption{The quality levels based on the annotations.}
\label{fig:quality-graph}
\end{figure}

We report the distribution of quality levels among the generated definitions in Figure \ref{fig:quality-graph}. After combining all three metrics into a unified quality level, we were able to show that more than 80\% of the generated definitions were meeting the quality level required for inclusion in a patient explainer form, while 20\% of the definitions were not. This is largely insufficient for this use case.

A large majority of the definitions which were not judged usable were nonetheless judged useful for machine learning purposes, with more than 96\% of definitions meeting the criteria for usefulness. 
For scoring or pretraining other models, e.g., in line with the recent work by \citet{Remy2022}, this appears 
sufficient to provide a strong signal.

However, around 3\% of the remaining definitions might turn out hurtful for the machine learning models, at least to some extent. Most of these definitions seem to concern less frequently used SnomedCT codes, however. This makes us confident that the dataset meets the requirements for usage for training retrieval models for the biomedical domain, but further work might still be required before this dataset can be used for other more critical use cases in biomedical NLP.

\section{Conclusion}
In this paper, we introduced a dataset of more than 400,000 computer-generated definitions for SnomedCT concepts, along with an quality control procedure applicable to biomedical definitions consisting of three axes (factuality, insight, and fluency) and a strict quality level classification based on these three axes. 

Our quality control demonstrated that this dataset is suitable to serve as a base for several biomedical pre-training tasks, for instance the development of robust biomedical retrieval models, and might act as a bronze standard for evaluating the inherent knowledge of biomedical concepts of large language models by rating the definitions they generate in the absence of a SnomedCT-sourced prompt. The usage of the definitions in user-facing scenarios is however not yet within reach.

\section*{Limitations}
The authors want to use the opportunity given by this column to highlight the fact that the definitions generated by this procedure do not all meet the standards required for presentation to users, or for reasoning-required scenarios, due to their imperfect quality. We release this dataset for building retreival-based systems, and evaluate large biomedical language models on the definition-generation task (and eventually for low-rank finetuning of existing language models).

\vspace{0.25cm}

In addition to the imperfect quality of the generated definitions and the presence of hurtful definitions in the dataset, it might also be useful to consider the bias induced by the choice of SnomedCT as our source of knowledge. While extensive, SnomedCT does not cover all possible relationships between concepts, and by biasing the output towards relationships present in SnomedCT, we might perpetuate existing biases in the data.

\vspace{0.25cm}

Another limitation is that we only evaluate the generated definitions on three metrics, but more could be relevant depending on the application. 

\vspace{0.25cm}

Finally, our rating of what is considered acceptable insight was biased towards what could possibly be condensed in short definitions (49 words on average), but longer definitions might sometimes be required to express the full range of nuance required by biomedical concepts. It is however difficult to estimate the value of omitted information.

\newpage
\section*{Ethics Statement}
The authors do not foresee any particular ethical concern about their work, as long as it is used within the guidelines outlined in the article. 

\vspace{0.25cm}

Releasing the dataset prevents unnecessary replications of this experiment, possibly with a less extensive QA than the one presented here.

\vspace{0.5cm}

\section*{Acknowledgements}
This work would not have been possible without the joint financial support of the Vlaams Agentschap Innoveren \& Ondernemen (VLAIO) and the RADar innovation center of the AZ Delta hospital group. 

\vspace{0.25cm}

Finally, I would also like to thank my co-supervisors, Kris Demuynck and Thomas Demeester, for their support and constructive advice during the ideation process, and all along the development of this project up to this very article.

\clearpage
\bibliography{anthology,custom}
\bibliographystyle{acl_natbib}

\clearpage
\appendix

\begin{figure*}[!t]
\section{Sample of generated definitions}
\label{sec:appendixA}

We provide an non-cherry-picked sample of the definitions found in our dataset.
These definitions are provided as an image instead of text, despite accessibility concerns, to reduce then chance these definitions get interpreted as authoritative by a machine learning model trained on scientific papers.

\vspace{0.5cm}
\includegraphics[width=1.0\textwidth]{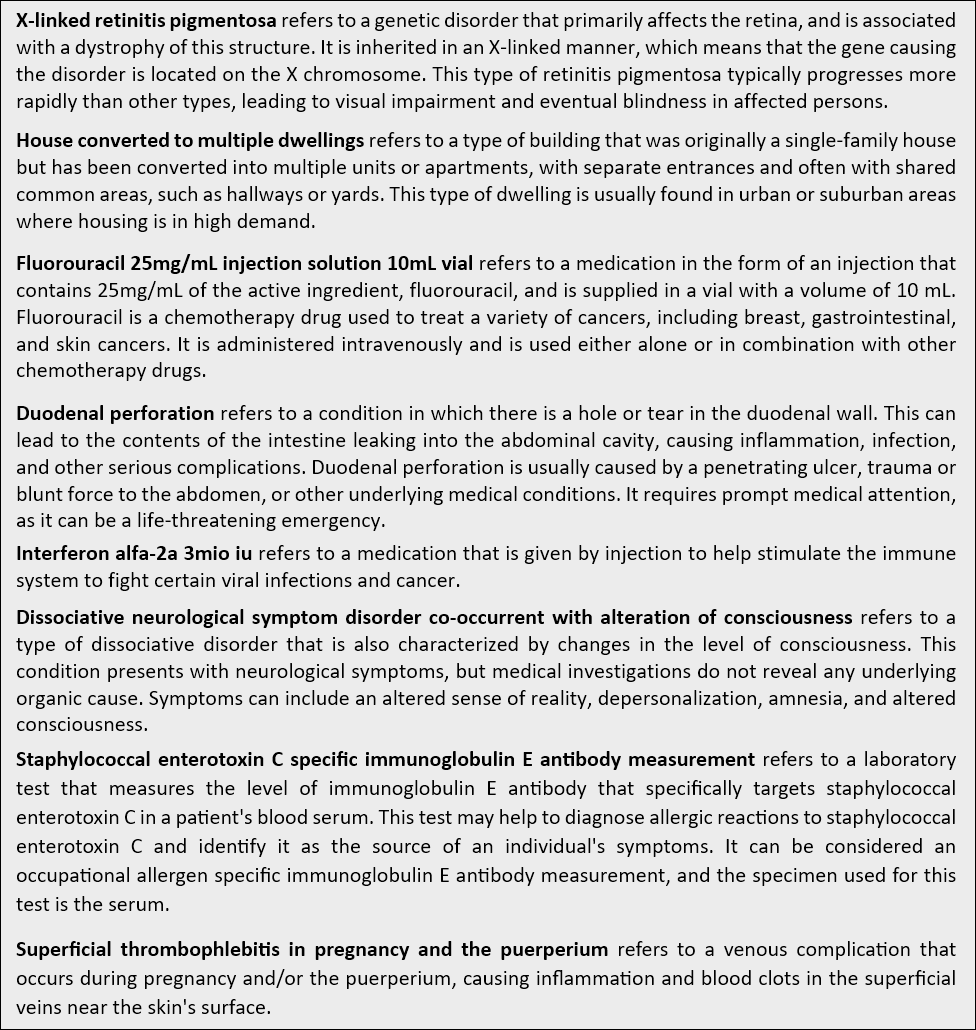}
\vspace{0.5cm}

\centerline{The complete set of definitions will be available for download once the dataset releases.}

\end{figure*}

\begin{figure*}

\section{Prompt and generation details}
\label{sec:appendixB}

While the entire code used to generate the definitions is also available in the supplementary materials, we wanted to provide in the paper a description of the prompt used to generate the definitions.

\vspace{0.25cm}

\lstset{
basicstyle=\small\ttfamily,
columns=flexible,
breaklines=true,
backgroundcolor=\color{gray!15},
frame=single,
framesep=10pt,
xleftmargin=10pt,
xrightmargin=10pt
}
\begin{lstlisting}
System: Assistant is a large language model, specialized in biomedical and clinical knowledge. It can answer questions about diseases, medications, and more. SnomedCT is a medical ontology, a standardized vocabulary of medical terms. It is reliable, and can be used to classify diseases and medications.

User: Let's talk about medical concepts. What does SnomedCT say about "{medical_concept}"?

Assistant: I found the following facts about "{medical_concept}" in SnomedCT:
{verbalized_snomed_facts}

User: Ok, thanks! Based on this, and your own medical knowledge, write a short definition of {medical_concept} in the style of MEDLINE or UMLS. Do not give a list of alternative names (also called) in the definition, the user already knows about them. Include some details about {required_details}. Leave out unimportant details if they are not useful inside a short definition. Start your reply immediately by the following words: Based on the given information, and my own medical knowledge, "{medical_concept}" refers to

Assistant:
\end{lstlisting}
\vspace{0.25cm}

We used greedy sampling, to get the most likely model output. We decided to remove from the dataset a small fraction of definitions where the model did not follow the template, or apologized for being unable to answer (1,131 concepts out of 423,201).

\vspace{1cm}
\section{Data availability}
Along with this paper, we release our entire dataset on HuggingFace at the following address:

\url{https://huggingface.co/datasets/FremyCompany/AGCT-Dataset}

\vspace{0.25cm}
The license for this work is subject to both SnomedCT and OpenAI API agreements. \\We strongly recommend checking those licenses before making use of this dataset.

\vspace{14cm}

\end{figure*}

\end{document}